\documentclass[conference]{IEEEtran}
\IEEEoverridecommandlockouts
\usepackage{cite}
\usepackage{amsmath,amssymb,amsfonts}
\usepackage{booktabs}
\usepackage{enumitem}
\usepackage{booktabs, tabularx}
\usepackage{multirow}
\usepackage{algorithm}
\usepackage{algpseudocode}
\usepackage{graphicx}
\usepackage{textcomp}
\usepackage{graphicx}
\usepackage[caption=false,font=footnotesize]{subfig} 
\usepackage{algorithm}
\usepackage{algpseudocode}
\usepackage{url}
\usepackage{xcolor}
\def\BibTeX{{\rm B\kern-.05em{\sc i\kern-.025em b}\kern-.08em
    T\kern-.1667em\lower.7ex\hbox{E}\kern-.125emX}}
\begin{document}

\title{User-Adaptive Meta-Learning for Cold-Start Medication Recommendation\\ with Uncertainty Filtering}

\author{
    \IEEEauthorblockN{Arya Hadizadeh Moghaddam\IEEEauthorrefmark{2}, Mohsen Nayebi Kerdabadi\IEEEauthorrefmark{2}, Dongjie Wang\IEEEauthorrefmark{2}, Mei Liu\IEEEauthorrefmark{3}, Zijun Yao\IEEEauthorrefmark{2}}
    \IEEEauthorblockA{\IEEEauthorrefmark{2}Electrical Engineering and Computer Science, University of Kansas
    \\\{a.hadizadehm, mohsen.nayebi, wangdongjie, zyao\}@ku.edu}
    \IEEEauthorblockA{\IEEEauthorrefmark{3} Clinical and Translational Science Institute, UF Health Science Center
    \\mei.liu@ufl.edu}
}

\maketitle

\begin{abstract}
Large-scale Electronic Health Record (EHR) databases have become indispensable in supporting clinical decision-making through data-driven treatment recommendations. However, existing medication recommender methods often struggle with a user (i.e., patient) cold-start problem, where recommendations for new patients are usually unreliable due to the lack of sufficient prescription history for patient profiling. 
While prior studies have utilized medical knowledge graphs to connect medication concepts through pharmacological or chemical relationships, these methods primarily focus on mitigating the item cold-start issue and fall short in providing personalized recommendations that adapt to individual patient characteristics. Meta-learning has shown promise in handling new users with sparse interactions in recommender systems. However, its application to EHRs remains underexplored due to the unique sequential structure of EHR data. To tackle these challenges, we propose MetaDrug, a multi-level, uncertainty-aware meta-learning framework designed to address the patient cold-start problem in medication recommendation. MetaDrug proposes a novel two-level meta-adaptation mechanism, including \textit{self-adaptation}, which adapts the model to new patients using their own medical events as support sets to capture temporal dependencies; and \textit{peer-adaptation}, which adapts the model using similar visits from peer patients to enrich new patient representations. Meanwhile, to further improve meta-adaptation outcomes, we introduce an uncertainty quantification module that ranks the support visits and filters out the unrelated information for adaptation consistency.
We evaluate our approach on the MIMIC-III and Acute Kidney Injury (AKI) datasets. Experimental results on both datasets demonstrate that MetaDrug consistently outperforms state-of-the-art medication recommendation methods on cold-start patients. 

\end{abstract}

\begin{IEEEkeywords}
Electronic Health Records, Meta-Learning, Medication Recommendation
\end{IEEEkeywords}

\section{Introduction}
\begin{figure}[t]
        \centering
	\includegraphics[width= 8.2cm]{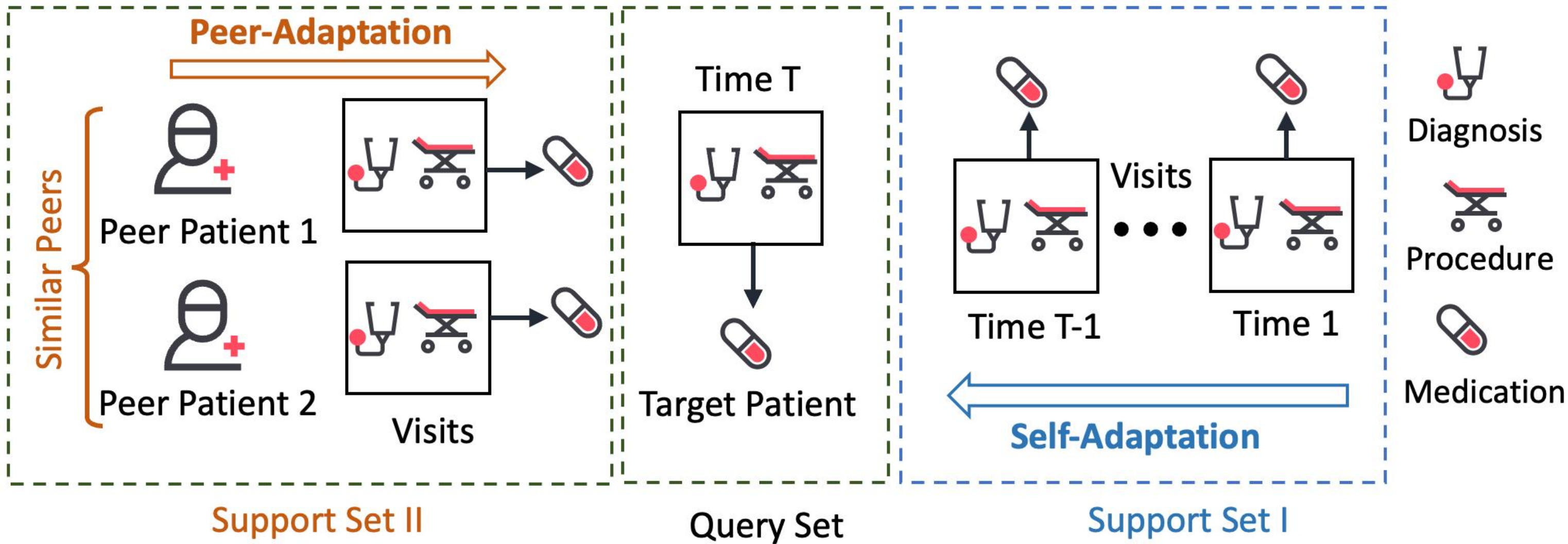}
	\caption{In MetaDrug, a two-level adaptation approach is employed to address the cold-start problem: (1) {self-adaptation}: the model adapts to a target patient using their own previous visits, and (2) peer-adaptation: the model further adapts by leveraging visits from peer patients with similar medical histories to the target patient.
    }
    \label{recommenderVSmedical}
    \vspace{-0.2cm}
\end{figure}

Electronic Health Records (EHRs) have been extensively utilized to support a variety of healthcare predictive tasks \cite{retain, jiang2023graphcare, theodorou2024consequence}. One of the primary applications is medication recommender systems \cite{vita, ARCI, molerec}, which have attracted significant interest from both clinical and research communities. These systems aim to recommend effective and safe medications for patients by analyzing their comprehensive medical history \cite{gamenet1}. 
To achieve accurate recommendations, it is necessary to learn robust patient representations from a series of discrete medical visits (e.g., hospital admissions), where each visit records various clinical events, including diagnosis, procedure, and medication. 
Accordingly, effectively capturing the sophisticated patient trajectory over a large number of visits is crucial for machine learning models to advance the performance of medication recommender systems \cite{bhoi2024refine}.

However, a significant issue for accurate medication recommendation is the user (i.e., patient) cold-start problem \cite{ye2021medpath,zhang2019metapred}. In healthcare recommender systems, this lack of patient history problem can occur at two levels: as a scarcity of patient visits or as a scarcity of medical codes in a patient’s history. While visit-based scarcity is commonly recognized, it does not always capture the whole extent of clinical information unavailability. A patient with numerous visits but a narrow range of medical codes also provides limited information for the model to learn from patients, whereas a patient with fewer visits but a broad set of codes can still offer a rich clinical context. From a collaborative filtering perspective, the number of medical codes shows how rich a patient profile is, making patients with limited code diversity particularly challenging to generate recommendations for. To address the cold-start issue, existing literature proposes to enrich item representation by incorporating domain-specific knowledge graphs, such as using ICD-9 or ATC ontology to classify observational diagnosis, procedures, and medications with hierarchical medical concepts \cite{tan2022metacare++, GRAM, SeqCare}. While these approaches indeed enrich item-level semantics, they overlook the possible enhancement of user profiling in tackling the patient cold-start problem.

Meta-learning has emerged as a promising approach in the literature to address the above cold-start problem in recommendation systems \cite{IJCAI}. Leveraging a few-shot adaptation technique, meta-learning treats each user as a distinct task, and utilizes their sparse interactions in history as a support set for rapid model adaptation \cite{MetaHIN,MELU,M2EU}.
However, applying meta-learning to cold-start medication recommendation has two unique challenges:

\begin{itemize}

\item The first challenge lies in how to effectively adapt a model to individual patients, given the unique characteristics of EHR data. Unlike conventional recommender systems, where meta-learning typically focuses on item-level adaptation, EHRs capture a sequence of medical visits; therefore, the adaptation needs to be applied on the visit level, where each visit involves multiple and co-occurring medical codes. Meanwhile, given the sparse nature of visits to cold-start patients, relying on self-owned visits alone is often insufficient. In this scenario, enriching the adaptation process by incorporating support visits from clinically similar patients is crucial to learn discriminative patient representations. 

\item The second challenge lies in how to filter out irrelevant support visits for effective adaptation. Existing meta-learning approaches often treat all the support user-item interactions equally during the adaptation phase, even when some support information does not reflect strong relevance.
To mitigate this disadvantageous impact, it is necessary to assess a relevance score for every support visit and exclude unrelated ones. Uncertainty Quantification (UQ) \cite{wang2023uncertainty, kweon2025uncertainty}, as a promising framework, can be fused in our work for this purpose. Based on the observation of UQ scores and the modeling of UQ prediction during the meta-training phase, we can infer UQ scores in the meta-testing phase to exclude visits with high uncertainty and further enhance the quality of user-specific adaptation.

\end{itemize}

To tackle these challenges, we propose MetaDrug, an uncertainty-aware meta-learning framework designed to tackle the patient cold-start problem in medication recommendation. As illustrated in Figure \ref{recommenderVSmedical}, MetaDrug enhances patient representation from two perspectives: self-adaptation and peer-adaptation, ensuring a comprehensive meta-learning process. Our contributions are as follows:
\begin{itemize}
    \item We propose a novel self-adaptation approach for meta-learning using EHR data. Our approach processes the adaptation of multiple medical codes in each visit simultaneously, better capturing the complexity of EHR visits. Over the sequence of visits, the model adapts incrementally and more effectively over time.
    \item We propose a peer-level adaptation strategy that identifies similar EHR visits from other peer patients. By performing an additional round of meta-adaptation with augmented similar patient experiences, our approach further mitigates data sparsity and enhances the representation of cold-start patients.
    \item We incorporate an uncertainty-based filtering mechanism that assigns a relevance score to each support visit based on its alignment with the patient's conditions. Irrelevant visits are subsequently filtered out during the meta-testing phase to enhance medication recommendation.
    \item We validate MetaDrug using the public MIMIC-III \cite{MIMIC3} dataset and a real-world AKI dataset. Our evaluation on both regular and cold-start patient groups demonstrates that MetaDrug outperforms state-of-the-art medication recommendation systems.
\end{itemize}

\section{Methodology}
\subsection{Problem Formulation}
The EHR data for patient $i$ is represented as a sequence of visits $\mathcal{D}_i = \{V_{i,t}\}_{t=1}^{T_i}$, where $V_{i,t}$ denotes the $t$-th visit (ordered chronologically), and $T_i$ is the total number of visits for patient $i$. For each visit $V_{i,t}$, a set of diagnosis or procedure codes is recorded, represented as $V_{i,t} = \{x_{j,t}\}_{j=1}^{|V_{i,t}|}$, where $|V_{i,t}|$ is the number of medical codes during visit $V_{i,t}$.


\noindent\textbf{Task:} 
Given a patient $i$ with a sequence of visits $\mathcal{D}_i = \{V_{i,t}\}_{t=1}^{T_i}$, the objective is to predict whether a medication is recommended or not at the last visit, denoted by 
$Y_{i,T_i} \in \{0,1\} ^ H$, where $H$ is the total number of candidate medications. Each entry $Y_{i,T_i}^{(j)}$ shows the presence or absence of the $j$-th medication being prescribed at visit $V_{i,T_i}$. The sequence of previously prescribed medications in the patient’s history, $\{Y_{i,t}\}_{t=1}^{T_{i}-1}$, serves as input for the meta-learning adaptation.
A patient is considered a cold-start if they fall below a certain percentile (e.g., the bottom 10\%) of the overall patient population in terms of their total number of recorded medical codes. The notation table is illustrated in Table \ref{table:notation}.


\begin{table*}[ht]
\small
\centering
\caption{Summary of Notations Used in the Methodology.}
\renewcommand{\arraystretch}{1.12}
\begin{tabularx}{\textwidth}{@{}lX lX@{}}
\toprule
\label{table:notation}
\textbf{Notation} & \textbf{Description} & \textbf{Notation} & \textbf{Description} \\
\midrule
$\mathcal{D}_i = \{V_{i,t}\}_{t=1}^{T_i}$ & EHR sequence for patient $i$. &
$V_{i,t} = \{x_{j,t}\}_{j=1}^{|V_{i,t}|}$ & Set of diagnosis/procedure codes in visit $t$. \\

$T_i$ & Total number of visits for patient $i$. &
$|V_{i,t}|$ & Number of medical codes in visit $V_{i,t}$. \\

$x_{j,t}$ & $j$-th medical code in visit $t$. &
$Y_{i,T_i} \in \{0,1\}^H$ & Multi-label vector of medications at last visit. \\

$H$ & Total number of candidate medications. &
$\hat{\boldsymbol{x}}_{j,t} \in \mathbb{R}^{d}$ & Embedding of medical code $x_{j,t}$. \\

$d$ & Hidden dimension size. &
$\hat{V}_{i,t} \in \mathbb{R}^{|V_{i,t}|\times d}$ & Embedding matrix for visit $t$. \\

$\hat{V}_{i}$ & Concatenated embedding matrix for all visits. &
$N$ & Maximum number of codes per patient. \\

$\boldsymbol{E}^{p}_i$ & Output of Patient-Transformer. &
$\hat{\boldsymbol{E}}^{p}_i \in \mathbb{R}^{d}$ & Averaged patient embedding. \\

$\boldsymbol{E}^{V_i}_{T_i}$ & Output of Visit-Transformer for last visit. &
$\hat{\boldsymbol{E}}^{V_i}_{T_i} \in \mathbb{R}^{d}$ & Averaged last-visit embedding. \\

$\boldsymbol{w}_g, \boldsymbol{b}_g$ & Parameters of gating layer. &
$\boldsymbol{g}_i$ & Gating vector for patient $i$. \\

$\boldsymbol{w}^i_1$ & Patient-specific predictive weight. &
$\boldsymbol{w}_1, \boldsymbol{w}_2$ & Weights of predictive layers. \\

$b_1, b_2$ & Bias terms of prediction layers. &
$\hat{y}_{i,T_i}$ & Predicted medication probabilities. \\

$\mathcal{L}_{(\cdot)}$ & Binary cross-entropy loss. &
$\alpha$ & Learning rate for adaptation.
 \\

$\theta$ &  Parameters of embedding \& transformer layers. & $\phi^{i}$ & Parameters of prediction \& gating layers.
 \\

$\mathcal{L}^{\text{self}}$ & Self-adaptation loss over prior visits. &
$\phi^{i}_1$ & Parameters after self-adaptation. \\

$\mathcal{L}^{\text{peer}}$ & Peer-adaptation loss using similar visits. &
$\phi^{i}_2$ & Parameters after peer-adaptation. \\

$\lambda$ & Number of top similar visits in peer-adaptation. &
$\mathrm{Jaccard}(V_{i,T_i}, V_{j,m})$ & Visit similarity based on shared codes. \\

$\mathcal{L}^{\text{query}}$ & Loss for global adaptation. &
$\mathcal{S}^{(j)}_{i,t_i}$ & Uncertainty score for medication $j$ at visit $t_i$. \\

$\hat{\mathcal{S}}_{i,t_i}$ & Predicted uncertainty score. &
$U^{i}_{t_i}$ & Averaged uncertainty score for visit $t_i$. \\

$\gamma$ & Threshold for uncertainty filtering. &
$\beta$ & Percentile to determine $\gamma$. \\

$f_{\text{slf-atn}}(\cdot)$ & Transformer-based UQ predictor. & & \\
\bottomrule
\end{tabularx}
\end{table*}

\subsection{Preliminary: Meta-Learning}

Meta-learning \cite{meta1, meta2, meta3, moghaddam2025meta}, referred to as \textit{learning to learn}, focuses on developing models that can rapidly adapt to new tasks with only a few examples. Unlike conventional machine learning, which optimizes model parameters based on a loss function, meta-learning optimizes for the ability to \textit{adapt} to quickly learn new tasks sampled from a distribution of related tasks. This approach captures higher-order knowledge that generalizes beyond specific datasets, which mimics how humans use prior experience to learn new skills efficiently.

Formally, meta-learning assumes a distribution over tasks $p(\mathcal{T})$, where each task $\mathcal{T}_i \sim p(\mathcal{T})$ contains a small \textit{support set} $\mathcal{Z}_i$ for local training and a \textit{query set} $\mathcal{Q}_i$ for meta-optimization. During the training phase, the model parameters $\Gamma$ are first adapted to each task via a few gradient-based updates on its support set to extract the task-specific parameters $\Gamma_i'$:
\begin{equation}
    \Gamma_i' = \Gamma - \alpha \nabla_\Gamma \mathcal{L}_{\mathcal{Z}_i}(f_\Gamma)
\end{equation}
where $\alpha$ denotes the learning rate for the inner adaptation. 
The model is then evaluated on the corresponding query set $\mathcal{Q}_i$, and the meta-learner updates the global parameters by minimizing the losses across all tasks:
\begin{equation}
    \Gamma \leftarrow \Gamma - \kappa \sum_i \nabla_\Gamma \mathcal{L}_{\mathcal{Q}_i}(f_{\Gamma_i'})
\end{equation}
where $\kappa$ represents the meta-learning rate for the outer update. 
This optimization process guides the model to learn a general initialization that enables it to quickly adapt to new tasks with only a few gradient updates.

The meta-learning framework can be categorized into three different groups: 
\textbf{(1) Metric-based methods}, which learn embedding spaces and similarity measures for few-shot inference \cite{chen2020variational}; 
\textbf{(2) Memory-based methods}, which augment neural architectures with external memory modules to store task-specific experience \cite{deng2020meta} and 
\textbf{(3) Optimization-based methods}, which directly learn a set of model parameters or an optimizer that enables fast adaptation \cite{finn2017model}. 
Among these, optimization-based approaches are widely adopted due to their generality and compatibility with gradient-based deep models.

Meta-learning treats each task as an independent learning episode, where the model learns not only the mapping between inputs and outputs but also how to be updated given limited test samples. By training across many such episodes, the model learns meta-knowledge that captures shared structure across tasks, such as common representations or learning dynamics. 
\subsection{Methodology Overview}
\begin{figure*}[ht]
        \centering
	\includegraphics[width= \textwidth]{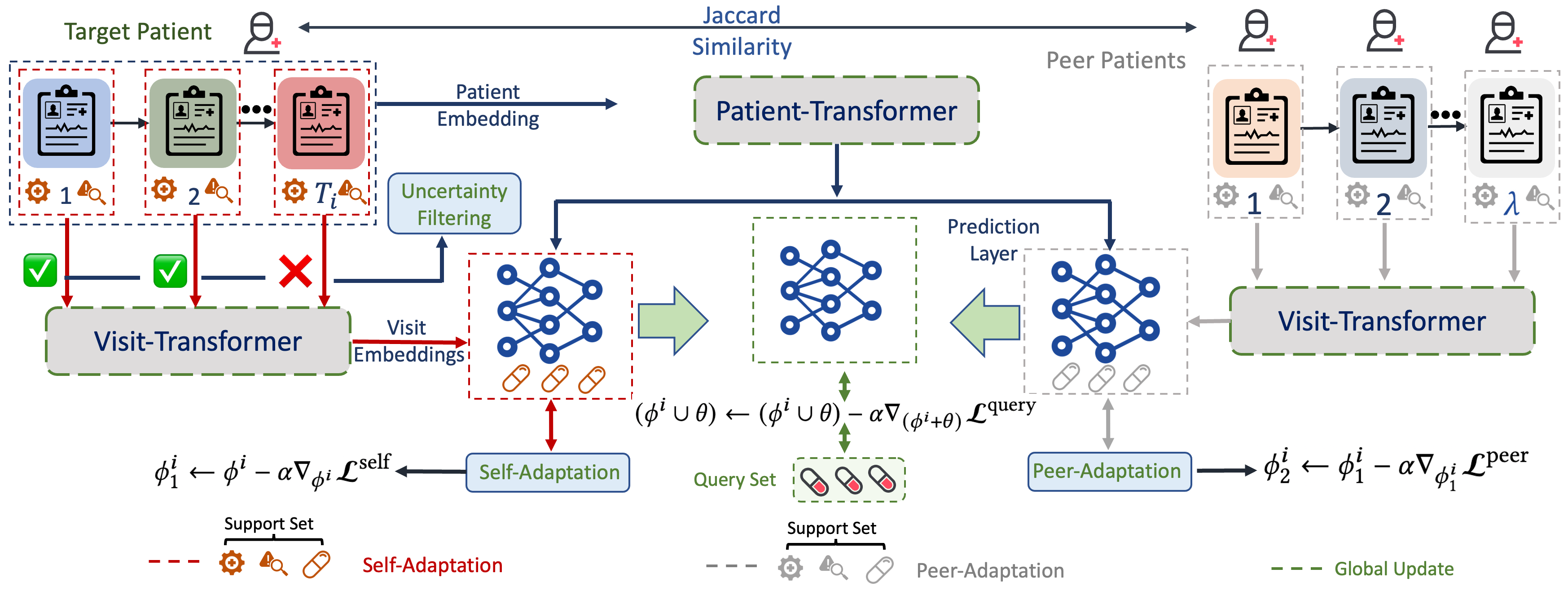}
    \caption{In the MetaDrug framework, patient embeddings are first generated using a Patient Transformer, while visit embeddings are obtained through a Visit Transformer. In the self-adaptation phase, the prediction layer is then adapted using the patient's historical visits, by a local update of the layer. During peer-adaptation, similar visits from other patients are identified based on Jaccard similarity, and the prediction layer undergoes a second round of local adaptation using these visits. The resulting adapted model is used to make predictions on the query set. Finally, the trained model is employed for Uncertainty Filtering, where a relevance score is computed for each visit. These scores are used to filter visits during meta-testing, thereby improving generalization performance.}
    
    \label{method}
\end{figure*}

As illustrated in Figure \ref{method}, MetaDrug comprises the following modules: (1) \textbf{Recommender Model:} This module consists of two Transformer encoders. The patient transformer generates patient embeddings from the patient's history, while the visit transformer encodes a visit with multiple medical codes. The prediction layer in this module utilizes a preference gating layer \cite{IJCAI} to dynamically adjust patient embedding for final medication prediction. 
(2) \textbf{Self-Adaptation:} This module adapts the recommender model on a target patient's historical visits using meta-learning and sequential modeling of those visits as support sets. 
(3) \textbf{Peer-Adaptation:} This process employs Jaccard similarity across EHR datasets by comparing the patient's most recent visit to visits of other patients. Therefore, even when the patient's prior visits contain limited items, the peer-adaptation mechanism can effectively compensate for cold-start sparsity.
(4) \textbf{Uncertainty Filtering:} A dedicated UQ predictive model is trained during meta-training and removes unrelated visits in meta-testing for better generalization in the self-adaptation phase.

\subsection{Recommender Model}
The Transformer \cite{vaswani2017attention} architecture has demonstrated its effectiveness in capturing the relationship among sequential inputs. Given that the structure of EHR encompasses two levels of relationship: the visit-level and code-level interdependencies, we employ two separate Transformer encoders to extract patient representations for medication recommendation.

At the beginning, we use an embedding layer to generate a representation for $j$-th medical code at $t$-th visit. The output is described in Equation \ref{eq:embd}, where $\hat{\boldsymbol{x}}_{j,t} \in \mathbb{R}^{d}$, and $d$ denotes the hidden dimension.
\begin{equation}
\label{eq:embd}
\hat{\boldsymbol{x}}_{j,t} = \text{Embedding}({\boldsymbol{x}_{j,t}})
\end{equation} 

\subsubsection{Patient-Transformer}\hspace{2pt}
We employ the first Transformer encoder that takes all medical codes recorded in a patient history to capture the interdependency among medical codes, aiming to extract meaningful code embeddings for patient-level processing.

We first construct code embedding matrix $\hat{V}_{i}$ by concatenating all the visit embedding matrices vertically as $\hat{V}_{i} = [\hat{V}_{i,1}, \hat{V}_{i,2}, \ldots, \hat{V}_{i,T_{i}}]$. Each visit is represented by stacking all the medical codes that are associated with it, denoted $\hat{V}_{i,t} = [\hat{x}_{1,t}, \hat{x}_{2,t}, \ldots, \hat{x}_{|V_{i,t}|,t}]$.
Eventually, we have all the code embeddings belonging to the patient $i$ as $\hat{V}_{i} \in \mathbb{R}^{N \times d}$, where $N$ denotes the maximum number of medical codes at the patient level, including padding for varying the number of codes between patients in a batch.


We then use $\hat{V}_{i}$ as input to the self-attention layer in patient-transformer, obtaining the output as defined in Equation \ref{eq:self_attention}:
\begin{equation}
\label{eq:self_attention}
\boldsymbol{E}^p_i = \text{Self-Attention}(\hat{{V}}_{i})
\end{equation} 

We average the embedding matrix $\boldsymbol{E}^p_i$ over all the codes to get the final output $\hat{\boldsymbol{E}}^p_i \in \mathbb{R}^{d}$ as patient embedding, which shows a comprehensive representation of the patient history.

\subsubsection{Visit-Transformer}\hspace{2pt}
In EHR analysis, a patient's visits provide valuable information about the progression of the patient's health condition. Among them, the most recent visit can be particularly significant as it captures the latest health profile \cite{retain}, therefore, it is directly relevant to the target medication.
To effectively extract and represent critical information from an individual visit, we utilize a second Transformer encoder specifically designed to generate embeddings for the target visit.

As illustrated in Equation \ref{eq:self_attention_2}, the input is the code embedding matrix of the last visit $\hat{{V}}_{i,T_{i}}$, and the output is $\boldsymbol{E}^{V_i}_{T_i}$: 
\begin{equation} 
\label{eq:self_attention_2} 
\boldsymbol{E}^{V_i}_{T_i} = \text{Self-Attention}(\hat{{V}}_{i,T_{i}}) 
\end{equation} 
where $\boldsymbol{E}^{V_i}_{T_i}$ represents the output of the encoding of all medical codes in the same visit, and $\hat{\boldsymbol{E}}^{V_i}_{T_i} \in \mathbb{R}^{d}$ represents the average embedding of $\boldsymbol{E}^{V_i}_{T_i}$, capturing the overall representation of the last visit, including information for all medical codes in the visits.

\subsubsection{Medication Recommendation}\hspace{2pt}
After obtaining the representation of patient $\hat{\boldsymbol{E}}^p_i \in \mathbb{R}^{d}$ and visit $\hat{\boldsymbol{E}}^{V_i}_{T_i} \in \mathbb{R}^{d}$ from the two Transformers, 
we first generate a personalized predictive layer weight inspired by the preference gating mechanism in \cite{IJCAI}, as shown in Equations \ref{eq:global_shared_knowledge} and \ref{eq:weight_u}.
Specifically, we use patient embedding $\hat{\boldsymbol{E}}^p_i$ as input to adjust the predictive parameters \( \boldsymbol{w}^{i}_1 \) as output customized to patient $i$ for final medication prediction, where $\boldsymbol{w}_g$, $\boldsymbol{b}_g$ are learnable gating parameters, diag($\boldsymbol{g}_i$) denotes the diagonal matrix whose diagonal entries are given by $\boldsymbol{g}_i$, and \(\boldsymbol{w}_1 \) is a learnable prediction parameter.
\begin{equation} 
\label{eq:global_shared_knowledge} 
\boldsymbol{g}_i = \sigma(\boldsymbol{w}_g \hat{\boldsymbol{E}}^p_i + b_g)
\end{equation} 
\begin{equation} 
\label{eq:weight_u} 
\boldsymbol{w}^i_1 = \text{diag}(\boldsymbol{g}_i){\boldsymbol{w}_1}
\end{equation} 

Next, we concatenate $\hat{\boldsymbol{E}}^p_i$ and  $\hat{\boldsymbol{E}}^{V_i}_{T_i}$ as input, and pass them through the first predictive layer consisting of $\boldsymbol{w}^i_1$ and \( {b}_1 \), followed by the second predictive layer consisting of $\boldsymbol{w}_2$ and \( {b}_2 \) as:
\begin{equation} 
\label{eq:weight_final} 
\hat{y}_{i,T_{i}} = \sigma(\boldsymbol{w}_2(\boldsymbol{w}^i_1 [\hat{\boldsymbol{E}}^p_i, \hat{\boldsymbol{E}}^{V_{i}}_{T_{i}}] + b_1) + b_2)
\end{equation}
where $\boldsymbol{w}^i_1\in \mathbb{R}^{d \times 2d}$ is patient-specific predictive weight obtained in Equation \ref{eq:weight_u}, $\boldsymbol{w}_2 \in \mathbb{R}^{H \times d}$, ${b}_1$, ${b}_2$ are learnable parameters for prediction, $H$ is the total number of candidate medications, and $\sigma(\cdot)$ indicates sigmoid function.

\subsubsection{Loss Function}\hspace{2pt}
For the loss function of medication prediction for patient $i$, we employ Equation \ref{eq:bce} to evaluate the Binary Cross Entropy (BCE) over every candidate medication, and aggregate them for the following meta-learning framework.
\begin{equation}
\label{eq:bce}
\begin{split}
\boldsymbol{\mathcal{L}}_{f\left(\hat{E}^p_i, \hat{E}^{V_{i}}_{T_{i}}, \theta, \phi^{i}\right), {Y}_{i,T_i}}=-\sum_{j=1}^{H} & {Y}^{(j)}_{i,T_i} \log \left(\hat{{y}}^{(j)}_{i,T_i} \right)\\
& +\left(1-{Y}^{(j)}_{i,T_i} \right) \log \left(1-\hat{{y}}^{(j)}_{i,T_i} \right)
\end{split}
\end{equation}
In the following, we define $\theta$ that contains the parameters of the embedding layer (Eq. \ref{eq:embd}), the patient transformer layer (Eq. \ref{eq:self_attention}), and the visit transformer layer (Eq. \ref{eq:self_attention_2}). The set of remaining parameters of the network, denoted as $\phi$, represents the prediction module, which includes two linear layers (Eq. \ref{eq:weight_final}) and the preference gating layers (Eqs. \ref{eq:global_shared_knowledge} and \ref{eq:weight_u}).

\subsection{Self-Adaptation}

To develop meta-learning for the patient cold-start problem in EHR, two unique challenges need to be addressed. First, unlike traditional recommendation systems, where each user interacts with a single item at a time, each patient interacts with a set of medical codes representing diagnoses, medications, or procedures in a visit. Second, effective EHR modeling must account for the sequential progression of multiple visits to accurately capture changes in a patient’s health status over time. 
These challenges necessitate an adaptation mechanism that can process each visit (in support set) sequentially and locally update the model. To address this, we propose a \emph{self-adaptation} approach that leverages the temporal structure of EHR visits, enabling the model to be adapted for more personalized medication recommendations.

In our approach, since the prediction layer makes the primary decision on medication recommendation based on patient representations,
we propose to perform local adaptation by updating the prediction layer (with parameter set denoted $\phi^{i}$) based on the support visits for each patient, meanwhile, the rest part of model (with parameter set denoted $\theta$) is frozen during local adaptation. This design ensures the stability of the adaptation process by preserving the generalized representations of patient and medical codes and reducing computational overhead. Additionally, this approach aligns with the core principle of meta-learning, which emphasizes updating task-specific parameters while maintaining the model's general-purpose capabilities \cite{MELU, M2EU}.

Formally, we first compute the loss function for each visit except for the last one (in query set) through Equation \ref{eq:loss_t}. Here, $\boldsymbol{\mathcal{L}}^{\text{self}}_t$ represents the loss value associated with the medications prescribed during the visit at time $t$. For simplicity, as $\theta$ is not changed, we remove it from the formulation. This loss is calculated based on the embeddings derived from the target patients $\hat{E}^p_i$ and the visits $\hat{E}_{t}^{V_i}$ at time $t$.
\begin{equation} 
\label{eq:loss_t} 
\boldsymbol{\mathcal{L}}^{\text{self}}_t = \boldsymbol{\mathcal{L}}_{f\left(\hat{E}^p_i, \hat{E}^{V_i}_{t},\phi^{i}\right), Y_{i,t}}
\end{equation} 

Next, we compute the overall meta-loss by averaging the local update losses over each visit, as defined in Equation \ref{eq:average}. Here, $\boldsymbol{\mathcal{L}}^{\text{self}}$ represents the loss value that will be used for local adaptation.

\begin{equation} 
\label{eq:average} 
\boldsymbol{\mathcal{L}}^{\text{self}} = \frac{1}{T_i - 1} \sum_{t=1}^{T_i - 1} \boldsymbol{\mathcal{L}}^{\text{self}}_t
\end{equation} 

Last, we update the prediction layer parameters based on the gradient of $\boldsymbol{\mathcal{L}}^{\text{self}}$ through Equation \ref{eq:phi}.

\begin{equation} 
\label{eq:phi} 
\phi^{i}_1 \leftarrow \phi^{i} - \alpha \nabla_{\phi^{i}} \boldsymbol{\mathcal{L}}^{\text{self}}
\end{equation} 

\begin{figure}[t]
        \centering
	\includegraphics[width= 0.35\textwidth]{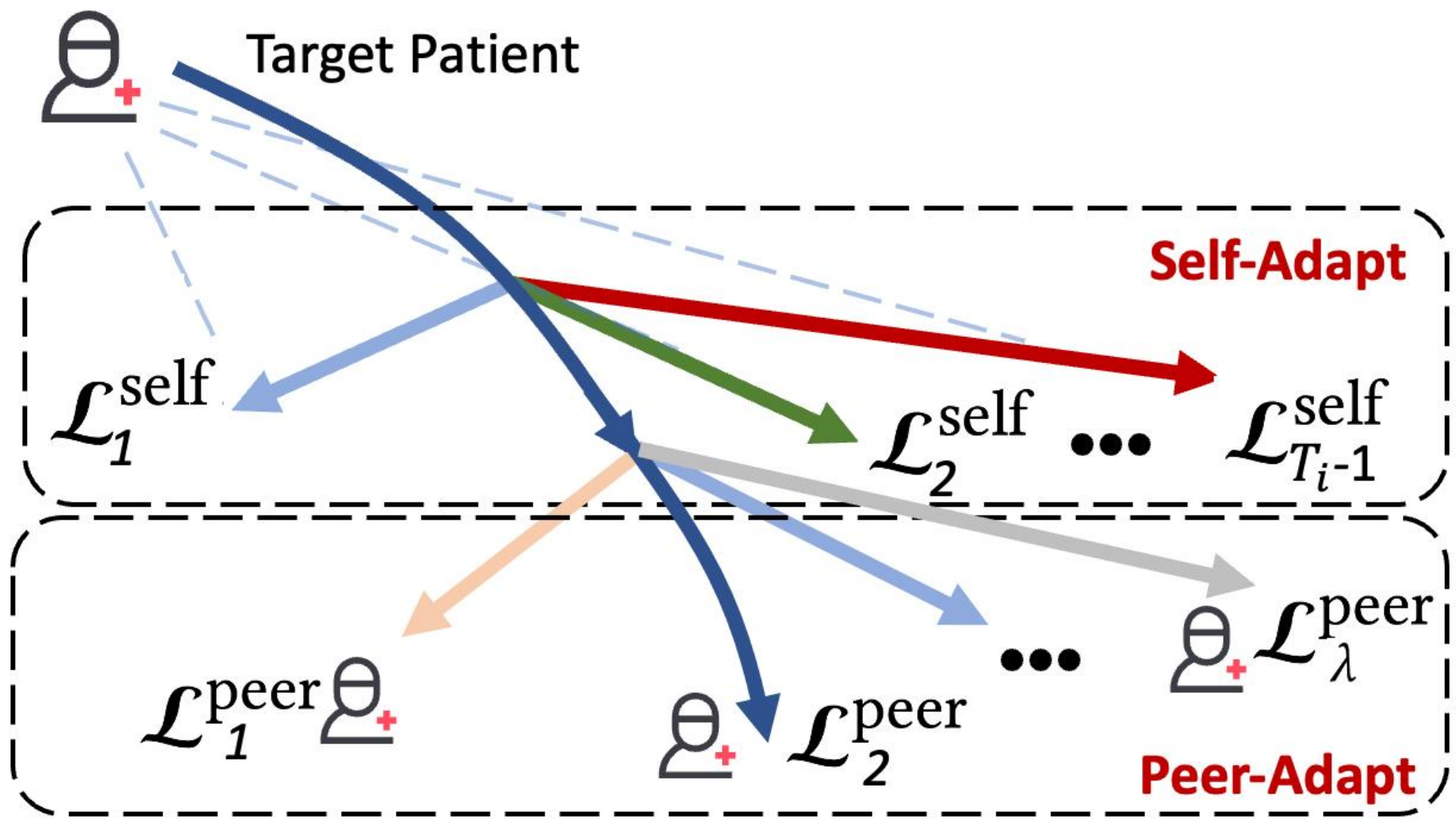}
    \caption{Self-adaptation and peer-adaptation for the target patient based on their previous visits and similar peers.}
    
    \label{visit_level}
\end{figure}

Through the self-adaptation step, the updated prediction layer has been personalized for the target patient, considering the previous visits and their associated medication codes. 

\subsection{Peer-Adaptation}
Self-adaptation is designed to sufficiently utilize the limited visits a patient has for model refinement in cold-start setting. However, given the nature of EHR data that patients frequently have a small number of visits, sometimes only one in the past, the data scarcity may still constrain the model’s ability to adapt for query set prediction. To overcome this limitation, we propose to identify similar visits from peer patients and locally update the prediction layer using those visits, utilizing inter-patient information for model adaptation, named \emph{peer-adaptation}.

In clinical settings, patients often have multiple visits over time, and therefore, defining patient similarity based on their entire medical history can be ineffective. A patient’s health status may change significantly across long histories, making earlier visits less relevant for current medication recommendations. Therefore, an effective strategy is to identify visits from other patients that share diagnosis and procedure codes with the target patient’s most recent visits. These shared codes are more likely to indicate similar clinical contexts, leading to more relevant medication suggestions and supporting a more effective adaptation phase.

Motivated by this, we leverage the set of diagnosis and procedure codes from the last visit $T_i$ of patient $i$ to identify similar visits among peer patients. To quantify similarity, we use the Jaccard similarity metric based on the experiments we have done using different similarity metrics. Based on this measure, we select the top $\lambda$ most similar visits from peer patients $j$ to support the adaptation process, as shown in Equation~\ref{eq:jaccardsearch}.
\begin{equation} 
\label{eq:jaccardsearch} 
\text{Similar}({V}_{i, T_i}) = \underset{{j \neq i, m \in \{1...T_j-1\}}}{\arg\max} \, \text{Jaccard}({V}_{i, T_i}, {V}_{j, m})
\end{equation} 

To adapt the recommendation model based on these similar visits, the first step is to compute the loss function using Equation \ref{eq:loss_t2}. In this equation, $n$ represents the $n^{\text{th}}$ similar visit up to a predefined number $\lambda$, $V_{j,m} \in \text{Similar}(V_{i, T_i})$ and $j$ is the peer patient, $k$ indicates the time of visit. 

\begin{equation} 
\label{eq:loss_t2} 
\boldsymbol{\mathcal{L}}^{\text{peer}}_n = \boldsymbol{\mathcal{L}}_{f\left(\hat{E}^p_i, \hat{E}^{ V_j}_{k},\phi^{i}_1\right), Y_{j,k}}
\end{equation} 

Next, we compute the overall loss value and update the model using Equations \ref{eq:average1} and \ref{eq:phi1}. The optimization process is guided by the weights derived from the self-adaptation, ensuring the model effectively incorporates the information from similar visits.
\begin{equation} 
\label{eq:average1} 
\boldsymbol{\mathcal{L}}^{\text{peer}} = \frac{1}{\lambda} \sum_{n=1}^{\lambda} \boldsymbol{\mathcal{L}}^{\text{peer}}_n
\end{equation} 
\begin{equation} 
\label{eq:phi1} 
\phi^{i}_2 \leftarrow \phi^{i}_1 - \alpha \nabla_{\phi_{1}^{i}} \boldsymbol{\mathcal{L}}^{\text{peer}}
\end{equation} 

As shown in Figure \ref{visit_level}, the optimized model now accounts for both the past visits of target patients and the similar visits of peer patients. This multi-level adaptation approach effectively addresses the cold-start problem by employing two support sets to ensure that patients with a limited number of medical codes can still be quickly and effectively adapted for medication recommendations.

\noindent \textbf{Global Adaptation}: Finally, after both self-adaptation and peer-adaptation, the global adaptation is conducted using Equations \ref{eq:loss_q} and \ref{eq:phi_query} for the patient:
\begin{equation} 
\label{eq:loss_q} 
\boldsymbol{\mathcal{L}}^{\text{query}} = \boldsymbol{\mathcal{L}}_{f\left(\hat{E}^p_i, \hat{E}^{V_i}_{T_i}, \theta, \phi_2^{i}\right), Y_{i,T_i}}
\end{equation} 
\begin{equation} 
\label{eq:phi_query} 
({\phi^{i}} \cup {\theta}) \leftarrow ({\phi^{i}} \cup {\theta}) - \alpha \nabla_{({\phi^{i}} \cup {\theta})} \boldsymbol{\mathcal{L}}^{\text{query}}
\end{equation}  
where the adjusted model is utilized for the query set predictions, and the parameters of the entire network containing the prediction layer $\phi^i$ and Transformers and embedding layers $\theta$ are optimized.

\begin{figure}[t]
        \centering
	\includegraphics[width= 0.48\textwidth]{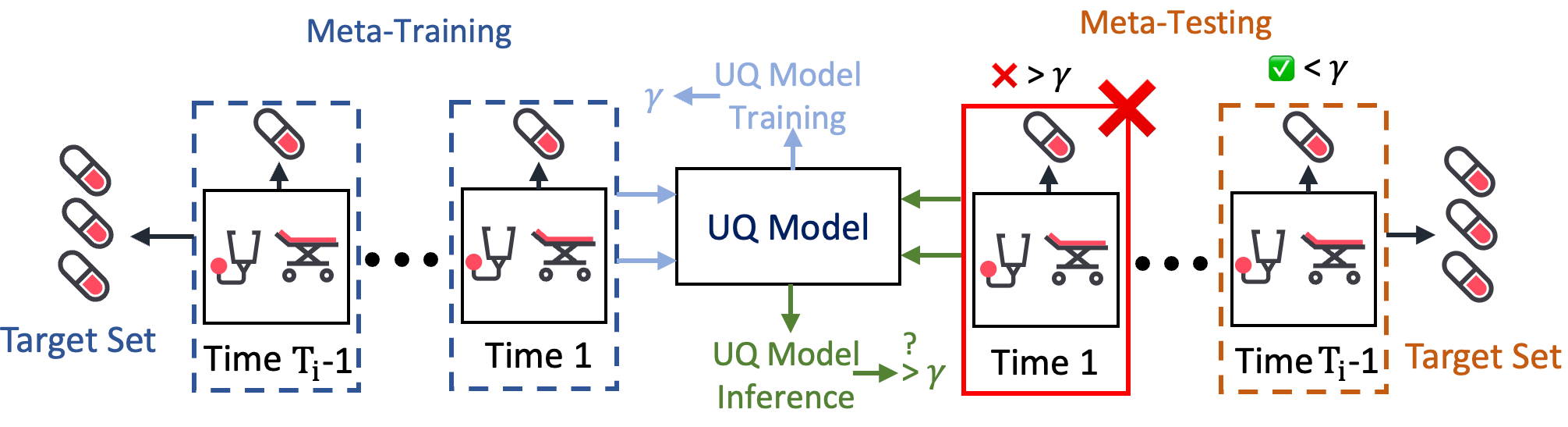}
    \caption{The uncertainty filtering model, trained on meta-training predictions, estimates uncertainty scores for each visit. During meta-testing, visits with high uncertainty are filtered out.}
    
    \label{uq_adapt}
\end{figure}
\subsection{Uncertainty Filtering (UF)}
Across a patient's medical history, some visits may not accurately represent their overall health profile, particularly those reflecting temporary conditions (e.g., a cold during a winter visit). Including such visits can mislead the self-adaptation process, as the model may attempt to adjust the patient-specific parameter $\phi^i$ based on irrelevant or non-informative data. Therefore, it is essential to identify and filter out these visits during inference.

As shown in Figure \ref{uq_adapt}, we propose an uncertainty quantification filtering mechanism. Specifically, we evaluate the model's confidence in predicting the medication of each support set visit using the patient embedding. If the uncertainty quantification \cite{gal2016dropout} measured exceeds a predefined threshold, we consider that visit to be uninformative and exclude it from the adaptation process.

To implement the uncertainty filtering approach, we first need to define an uncertainty quantification method. Based on our experiments, we chose \emph{Auxiliary Error Prediction} \cite{thiagarajan2020building} which formulates the uncertainty score based on the difference between the actual label and the predicted probability for visit $t$ and medication $j$, defined as ${\mathcal{S}}^{(j)}_{i,t_i} = \left| Y^{(j)}_{i,t_i} - \hat{y}^{(j)}_{i,t_i} \right|.$
In evaluation, we empirically validate more UQ formulations, including \emph{NC Dropout}, and \emph{Deep Ensemble}. After selecting a quantification method, we employ a transformer-based model that directly maps the patient embedding $\hat{\boldsymbol{E}}^{p}_{i}$ to a predicted uncertainty score using Equation \ref{eq:uq_prediction} for each visit. 
\begin{equation} 
\label{eq:uq_prediction} 
{\hat{\mathcal{S}}}_{i,t_i} = f_{\text{slf-atn}}(\hat{\boldsymbol{E}}^{p}_{i})
\end{equation}
Once the uncertainty model outputs an uncertainty value for each predicted medication in the support set, we proceed to optimize only the uncertainty quantification model.

Next, we compute an overall uncertainty score for each patient visit by averaging the uncertainty values corresponding to predicted positive labels using Equation \ref{eq:u_value}, where $U^{i}_{t_i}$ equals the uncertainty score for patient $i$ at visit $t_i$. 
\begin{equation} 
\label{eq:u_value} 
U^{i}_{t_i} = \operatorname{Avg}\left( \boldsymbol{\mathcal{S}}^{(j)}_{i, t_i} \;\middle|\; \hat{\boldsymbol{y}}^{(j)}_{i, t_i} > 0.5 \right)
\end{equation}

We then collect all such uncertainty scores $\{U^{i}_{t_i} | t_i < T_i \}$ across the training set and sort them to determine a threshold $\gamma$, that differentiates the top $\beta$ percent of uncertainty values from others. This threshold represents the uncertainty cutoff used to filter out noisy or misleading visits. We retain only the patient's visits with uncertainty scores below the $\gamma$ during meta-testing. This filtering aims to improve the self-adaptation process by ensuring that only reliable, informative visits contribute to the personalized adaptation of the model. 

The overall process of the proposed meta-learning framework is outlined in Algorithm 1, as it illustrates how two levels of local adaptations are performed (lines 5-7), how the adapted prediction layer is further utilized for the global update (line 8), how the UQ value $\gamma$ is achieved (lines 11) in the training phase, and how the model filters unrelated visits for each patient and perform adaptation (lines 12-13).

\begin{algorithm}[t]
\caption{MetaDrug Algorithm}
\label{algo:1}
\begin{algorithmic}[1]
\Require Patients from dataset $\mathcal{D}$, number of similar patients for peer adaptation $\lambda$, and learning rate $\alpha$.
\Ensure \textbf{Meta-Training Phase:} Trained parameters $\phi^i \cup \theta$, and uncertainty threshold $\gamma$.
\State Initialize encoder layers $\theta$ and prediction layer $\phi^i$.
\While{not converged}
    \State Sample batch $\mathcal{B}$ of patients from $\mathcal{D}$.
    \For{each patient $i \in \mathcal{B}$}
        \State Obtain embeddings and adapt the model using Eqs.~\ref{eq:embd} to \ref{eq:weight_final}.
        \State Compute BCE loss using Eqs.~\ref{eq:bce} to \ref{eq:average}.
        \State Identify similar visits (Eq.~13), then compute $\phi^{i}_2$ using Eqs.~\ref{eq:loss_t2}, \ref{eq:average1}, and \ref{eq:phi1}.
        \State Update $({\phi^{i}} \cup {\theta})$ using Eqs.~\ref{eq:loss_q} and \ref{eq:phi_query}.
    \EndFor
\EndWhile
\State Train UQ model to obtain $\gamma$ using Eqs.~\ref{eq:uq_prediction} and \ref{eq:u_value}.
\vspace{0.5em}
\Require A sample patient $D_t$ from the test set.
\Ensure \textbf{Meta-Testing Phase:} Recommended medications.
\State Filter out visits with UQ values above $\gamma$.
\State Perform self and peer adaptation similar to lines 5 to 7.
\State Recommend medications using Eq.~\ref{eq:weight_final}.
\end{algorithmic}
\end{algorithm}
\section{Evaluation}

\begin{table}[t]
\centering
\caption{Statistics of the datasets}
\label{Statistical}
\makebox[\linewidth]{
    \begin{tabular}{@{}lcc@{}}
    \hline
    Metric & \multicolumn{1}{c}{\bf{MIMIC-III}} & \multicolumn{1}{c}{\bf{AKI}}\\
        \hline
    \# of patients & 6514 & 24881  \\
    Avg. \# of drugs per patient  & 48.13 & 72.38  \\
    Avg. \# of diagnosis per patient  & 37.26  & 111.31  \\
    Avg. \# of procedures per patient  & 10.641 & 133.95  \\

    Avg \# of target drugs per patient  & 28.77 & 26.73  \\
    \# of visits per patient        & 2.66 & 3.65  \\
    
    \hline
    \end{tabular}
}
\end{table}

\begin{table*}[ht]
\centering
\caption{Performance comparison between MetaDrug and other baseline models on the MIMIC-III dataset. Values show mean $\pm$ 95\% CI across folds.}
\renewcommand{\arraystretch}{1.5}
\setlength{\tabcolsep}{2.5em}
\begin{tabular}{lcccc}
\toprule
\textbf{Models} & \textbf{PRAUC}$\uparrow$ & \textbf{F1}$\uparrow$ & \textbf{Jaccard}$\uparrow$ & \textbf{DDI}$\downarrow$ \\
\midrule
Deepr            & 0.723 $\pm$ 0.004 & 0.610 $\pm$ 0.005 & 0.450 $\pm$ 0.003 & 0.073 $\pm$ 0.002 \\
GameNet          & 0.694 $\pm$ 0.003 & 0.592 $\pm$ 0.005 & 0.432 $\pm$ 0.005 & 0.068 $\pm$ 0.002 \\
Micron           & 0.734 $\pm$ 0.004 & 0.623 $\pm$ 0.005 & 0.466 $\pm$ 0.003 & \textbf{0.066} $\pm$ 0.003 \\
Retain           & 0.727 $\pm$ 0.002 & 0.609 $\pm$ 0.003 & 0.451 $\pm$ 0.003 & 0.074 $\pm$ 0.004 \\
MELU             & 0.657 $\pm$ 0.008 & 0.566 $\pm$ 0.006 & 0.408 $\pm$ 0.007 & 0.071 $\pm$ 0.003 \\
SafeDrug         & 0.683 $\pm$ 0.006 & 0.584 $\pm$ 0.005 & 0.425 $\pm$ 0.003 & 0.067 $\pm$ 0.002 \\
Transformer      & 0.702 $\pm$ 0.005 & 0.597 $\pm$ 0.004 & 0.440 $\pm$ 0.006 & 0.081 $\pm$ 0.004 \\
MoleRec          & 0.703 $\pm$ 0.009 & 0.592 $\pm$ 0.008 & 0.432 $\pm$ 0.008 & 0.069 $\pm$ 0.003 \\
ARCI             & 0.722 $\pm$ 0.010 & 0.608 $\pm$ 0.009 & 0.449 $\pm$ 0.011 & 0.076 $\pm$ 0.003 \\
\textbf{MetaDrug} & \textbf{0.752} $\pm$ 0.011 & \textbf{0.641} $\pm$ 0.009 & \textbf{0.494} $\pm$ 0.008 & 0.083 $\pm$ 0.004 \\
\bottomrule
\label{tb:perf_mimic}
\end{tabular}
\vspace{-0.3cm}
\end{table*}

\begin{table*}[ht]
\centering
\caption{Performance comparison between MetaDrug and other baseline models on the AKI dataset. Values show mean $\pm$ 95\% CI across folds.}
\renewcommand{\arraystretch}{1.5}
\setlength{\tabcolsep}{2.5em}
\begin{tabular}{lcccc}
\toprule
\textbf{Models} & \textbf{PRAUC}$\uparrow$ & \textbf{F1}$\uparrow$ & \textbf{Jaccard}$\uparrow$ & \textbf{DDI}$\downarrow$ \\
\midrule
Deepr            & 0.508 $\pm$ 0.003 & 0.372 $\pm$ 0.004 & 0.298 $\pm$ 0.003 & 0.071 $\pm$ 0.002 \\
GameNet          & 0.489 $\pm$ 0.003 & 0.357 $\pm$ 0.003 & 0.286 $\pm$ 0.002 & 0.072 $\pm$ 0.003 \\
Micron           & 0.472 $\pm$ 0.002 & 0.339 $\pm$ 0.003 & 0.274 $\pm$ 0.002 & \textbf{0.064} $\pm$ 0.002 \\
Retain           & 0.491 $\pm$ 0.003 & 0.354 $\pm$ 0.004 & 0.282 $\pm$ 0.004 & 0.085 $\pm$ 0.002 \\
MELU             & 0.402 $\pm$ 0.002 & 0.281 $\pm$ 0.004 & 0.216 $\pm$ 0.005 & 0.079 $\pm$ 0.002 \\
SafeDrug         & 0.474 $\pm$ 0.002 & 0.348 $\pm$ 0.002 & 0.271 $\pm$ 0.003 & 0.078 $\pm$ 0.003 \\
Transformer      & 0.483 $\pm$ 0.003 & 0.352 $\pm$ 0.002 & 0.275 $\pm$ 0.003 & 0.082 $\pm$ 0.002 \\
MoleRec          & 0.498 $\pm$ 0.002 & 0.361 $\pm$ 0.003 & 0.288 $\pm$ 0.007 & 0.073 $\pm$ 0.002 \\
ARCI             & 0.502 $\pm$ 0.004 & 0.369 $\pm$ 0.003 & 0.301 $\pm$ 0.003 & 0.080 $\pm$ 0.002 \\
\textbf{MetaDrug} & \textbf{0.533} $\pm$ 0.005 & \textbf{0.390} $\pm$ 0.004 & \textbf{0.333} $\pm$ 0.003 & 0.086 $\pm$ 0.005 \\
\bottomrule
\label{tb:perf_aki}
\end{tabular}
\vspace{-0.3cm}
\end{table*}

\subsection{Datasets}
In this research, two datasets are utilized for training and evaluation:
\begin{itemize}
    \item \textbf{MIMIC-III:} MIMIC-III \cite{MIMIC3} is an open-access dataset containing data for over 40,000 patients admitted to the CCU between 2001 and 2002. For this study, we selected patients with more than one prior visit to enable meta-learning adaptation. All available visits for these patients were used to train the model.

    \item \textbf{AKI:} The Acute Kidney Injury (AKI) dataset \cite{liu2022development} was collected from the University of Kansas Medical Center between 2009 and 2021, and consists of over 135,000 hospitalized patients at risk of AKI. We include 50 percent of the patients from the AKI dataset.
\end{itemize}

\subsection{Experimental Setup}
In this study, we focus on recommending prescriptions for the most recent visit, utilizing conditions and procedures up to visit  $T_i$ to predict the target medication at $T_i$. To ensure sufficient data for meta-adaptation, we excluded patients with only one visit. 
We employed ATC level 3 classification for prescriptions in both datasets. For MetaDrug and the baseline models, we adopted a greedy approach to identify the hyperparameters that perform the best. Once the optimal hyperparameters were determined, we evaluated the models using an 80\% training and 20\% testing split, performing five independent folds to report the confidence intervals of the results. For the proposed method, the optimal configuration includes: $\alpha = 0.01$, $\lambda = 3$, hidden dimension $d = 256 $, $\beta = 20$, and the number of heads set to 1. The dataset statistics are presented in Table \ref{Statistical}. The implementation code is available online\footnote{https://github.com/aryahm1375/MetaDrug}. 

\subsection{Evaluation Metrics}
To evaluate MetaDrug, we employ classification metrics used for multi-label classification tasks. The metrics used include Jaccard, F1, and PRAUC. Below, the formulations for these metrics are provided:
\textbf{Jaccard} measures the similarity between two sets as the ratio of their intersection to their union, relative to the ground truth.
\begin{equation}
\label{eq:jaccard}
\textbf { Jaccard }=\frac{1}{\sum_i^N 1} \sum_i^N \frac{\left|\boldsymbol{Y}_{i, T_i} \cap \hat{\boldsymbol{Y}}_{i,T_{i}}\right|}{\left|\boldsymbol{Y}_{i, T_i} \cup \hat{\boldsymbol{Y}}_{i,T_{i}}\right|}
\end{equation} 
where, $N$ equals the number of patients and $\hat{\boldsymbol{Y}}_{i,T_{i}} = \{ \hat{\boldsymbol{y}}_{i,T_{i}} > \eta \}$ where $\eta$ shows a threshold, selected using a greedy search approach from values from 0 to 1. 0.3 is chosen with the best performance.

\noindent \textbf{F1} score is based on precision and recall, where $Recall_i = |\boldsymbol{Y}_{i, T_i} \cap \hat{\boldsymbol{Y}}_{i,T_{i}}| / \boldsymbol{Y}_{i, T_i}|$ and $Precision_i = |\boldsymbol{Y}_{i, T_i} \cap \hat{\boldsymbol{Y}}_{i,T_{i}}| / |\hat{\boldsymbol{Y}}_{i,T_{i}}|$. As shown in Equation \ref{eq:f1}, the final score is achieved.
\begin{equation}
\label{eq:f1}
\textbf{F1}=\frac{1}{\sum_i^N 1} \sum_i^N \frac{2 \times \text{Precision}_i \times \text{Recall}_i}{\text{Precision}_i + \text{Recall}_i}
\end{equation} 

\noindent  \textbf{PRAUC} shows the performance of a classification model with the area under the precision-recall curve. The formulation is shown in Equation \ref{eq:prauc_final}, where  $\operatorname{PRAUC}^i=\sum_{k=1}^{|Y|} \operatorname{Precision}(k)^i \Delta \operatorname{Recall}^i$ and $\Delta \operatorname{Recall}^i=\operatorname{Recall}(k)^i - \hspace{2pt} \operatorname{ Recall}(k-1)^i$, and $k$ is the sequence rank. 
\begin{equation}
\label{eq:prauc_final}
\operatorname{\textbf{PRAUC}}= \frac{1}{\Sigma_i^N 1} \sum_i^N \operatorname{PRAUC}^i
\end{equation}

\noindent \textbf{DDI} rate assesses medication safety by evaluating the frequency of interactions among predicted prescriptions \cite{gamenet1}. It is defined in Equation \ref{DDI}, where $m_l$ represents the predicted output of the $l^{th}$ prescription, and $G$ denotes the DDI graph.
\begin{equation}
\label{DDI}
\textbf {DDI }=\frac{\sum_i^N \sum_{l, j}\left|\left\{\left(m_l, m_j\right) \in \hat{\boldsymbol{Y}}_{i,T_{i}} \mid\left(m_l, m_j\right) \in G_d\right\}\right|}{\sum_i^N \sum_{l, j} 1}
\end{equation}

\begin{table*}[t]
\centering
\caption{Ablation study on the contribution of each module on the MIMIC-III dataset. Values show mean $\pm$ 95\% confidence intervals across folds.}
\renewcommand{\arraystretch}{1.5}
\setlength{\tabcolsep}{2.5em}
\begin{tabular}{lcccc}
\toprule
\multicolumn{5}{c}{\textbf{MIMIC-III}} \\
\midrule
\textbf{Models} & \textbf{PRAUC}$\uparrow$ & \textbf{F1}$\uparrow$ & \textbf{Jaccard}$\uparrow$ & \textbf{DDI}$\downarrow$ \\
\midrule
\textbf{MetaDrug}       & \textbf{0.752} $\pm$ 0.011 & \textbf{0.641} $\pm$ 0.009 & \textbf{0.494} $\pm$ 0.008 & 0.083 $\pm$ 0.004 \\
MetaDrug w/o UF         & 0.746 $\pm$ 0.008 & 0.633 $\pm$ 0.007 & 0.489 $\pm$ 0.007 & 0.077 $\pm$ 0.003 \\
Peer-Adaptation         & 0.728 $\pm$ 0.007 & 0.615 $\pm$ 0.006 & 0.458 $\pm$ 0.007 & 0.074 $\pm$ 0.004 \\
Self-Adaptation         & 0.740 $\pm$ 0.009 & 0.630 $\pm$ 0.009 & 0.480 $\pm$ 0.008 & 0.076 $\pm$ 0.003 \\
No Adaptation           & 0.714 $\pm$ 0.012 & 0.597 $\pm$ 0.009 & 0.440 $\pm$ 0.009 & \textbf{0.072} $\pm$ 0.002 \\
\bottomrule
\end{tabular}

\label{tb:ablation_mimic_consistent}
\vspace{-0.3cm}
\end{table*}

\begin{table*}[t]
\centering
\caption{Ablation study on the contribution of each module on the AKI dataset. Values show mean $\pm$ 95\% confidence intervals across folds.}
\renewcommand{\arraystretch}{1.5}
\setlength{\tabcolsep}{2.5em}
\begin{tabular}{lcccc}
\toprule
\multicolumn{5}{c}{\textbf{AKI}} \\
\midrule
\textbf{Models} & \textbf{PRAUC}$\uparrow$ & \textbf{F1}$\uparrow$ & \textbf{Jaccard}$\uparrow$ & \textbf{DDI}$\downarrow$ \\
\midrule
\textbf{MetaDrug}       & \textbf{0.533} $\pm$ 0.005 & \textbf{0.390} $\pm$ 0.004 & \textbf{0.333} $\pm$ 0.003 & 0.086 $\pm$ 0.005 \\
MetaDrug w/o UF         & 0.527 $\pm$ 0.003 & 0.384 $\pm$ 0.002 & 0.325 $\pm$ 0.002 & 0.082 $\pm$ 0.003 \\
Peer-Adaptation         & 0.504 $\pm$ 0.003 & 0.356 $\pm$ 0.003 & 0.304 $\pm$ 0.002 & \textbf{0.078} $\pm$ 0.002 \\
Self-Adaptation         & 0.519 $\pm$ 0.002 & 0.377 $\pm$ 0.003 & 0.315 $\pm$ 0.004 & 0.080 $\pm$ 0.003 \\
No Adaptation           & 0.493 $\pm$ 0.004 & 0.346 $\pm$ 0.003 & 0.293 $\pm$ 0.003 & 0.084 $\pm$ 0.002 \\
\bottomrule
\end{tabular}
\label{tb:ablation_aki_consistent}
\vspace{-0.3cm}
\end{table*}

\subsection{Baselines}
We utilize state-of-the-art healthcare predictive methods and prescription recommendation systems as baselines: 
\begin{itemize}
\item \textbf{Deepr} \cite{nguyen2016mathtt} an end-to-end model, which transforms features into sequences of discrete elements and uses a convolutional neural network to detect patterns.
\item \textbf{GAMENet} \cite{gamenet1} integrates a memory module and graph neural networks to utilize the drug-drug interaction knowledge graph. \item \textbf{Micron} \cite{yang2021change} is a drug recommendation model based on residual recurrent neural networks, designed for patient health change. 
\item \textbf{Retain} \cite{retain} leverages a dual-RNN framework to capture the interpretable impact of medical visits and features. 
\item \textbf{MELU} \cite{MELU} is a meta-learning-based recommender system designed to address the cold-start problem as the first paper offering a meta-learning-based solution. \item \textbf{SafeDrug} \cite{safedrug} employs a global message-passing neural network to encode the functionality of prescription molecules while considering DDI to recommend safe prescriptions. \item \textbf{Transformer} \cite{vaswani2017attention} is an important baseline for comparison in our research, as our encoder is mainly based on self-attention. 
\item \textbf{MoleRec} \cite{molerec} introduces a structure-aware encoding approach with a hierarchical architecture to model interactions while addressing DDI. 
\item \textbf{ARCI} \cite{ARCI} uses a multilevel transformer approach to connect drugs from different visits while considering the medical intentions.
\end{itemize}

\subsection{Results and Discussion}
In the following, these research questions are addressed:
\begin{itemize}
\item \textbf{RQ1} How does the performance of MetaDrug compare to other state-of-the-art recommendation methods and health informatics models?

\item \textbf{RQ2} What is the contribution of each level of adaptation and uncertainty filtering to the overall performance?

\item \textbf{RQ3} How does MetaDrug perform in recommending medications for cold-start patients?

\item \textbf{RQ4:} How do various uncertainty quantification methods affect the performance of MetaDrug? 

\end{itemize}

\subsubsection{RQ1: Overall Performance Comparison with Baselines}

\hspace{5pt} As shown in Tables \ref{tb:perf_mimic} and \ref{tb:perf_aki}, MetaDrug outperforms all previous baselines across the classification metrics. The performance differences between MetaDrug and other medication recommendation models, such as SafeDrug, MoleRec, ARCI, Micron, and GameNet, highlight that even with a relatively simple encoder, the meta-adaptation framework delivers superior results compared to models with more sophisticated encoders. The reason is that medical datasets often involve a limited number of patients or have sparse patient-medical code interaction, and meta-learning shows enhancement in these types of datasets. For time-series analysis, ARCI employs a two-level Transformer-based approach for feature extraction, and Retain incorporates a multi-level attention mechanism. However, despite their complexity, these models fail to match the performance of MetaDrug. This result underscores that incorporating the time-series analysis into the meta-learning framework enhances the recommendation performance. Moreover, the comparison between MetaDrug and MELU demonstrates that self-adaptation is more effective within the EHR framework than the item-level adaptation typically employed in recommendation systems. Finally, the comparison between the Transformer baseline and MetaDrug confirms that combining a patient and visit-transformer with meta-learning significantly improves the overall performance of the normal Transformer model. Although the DDI rate of the proposed method is higher than that of models incorporating DDI loss, such as SafeDrug, GameNet, MoleRec, and Micron, the marginal differences in classification metrics indicate that our method remains effective for medication recommendation, even with a higher DDI.

\begin{figure*}[t]
        \centering
	\includegraphics[width= 0.98\textwidth]{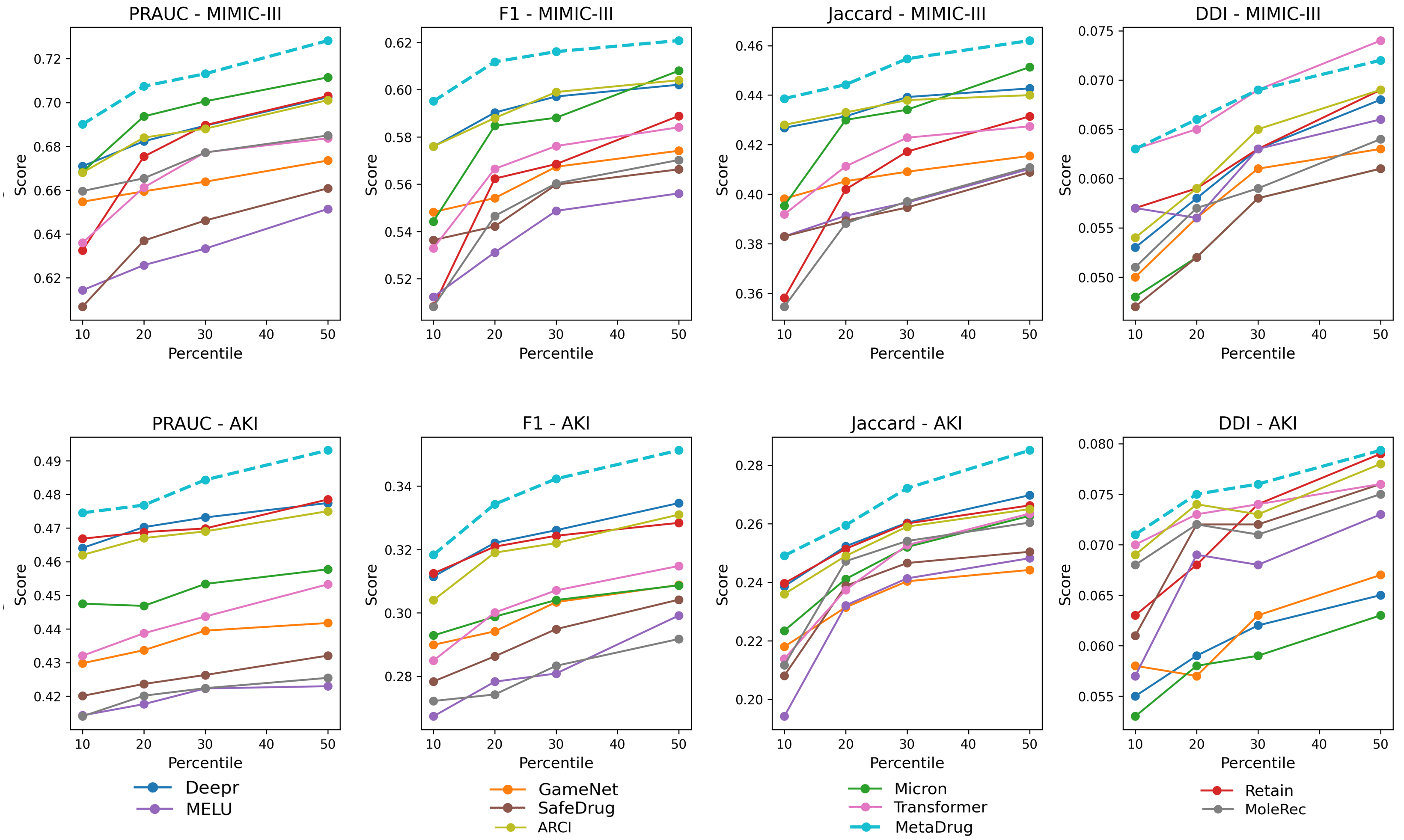}
    \caption{Performance comparison of MetaDrug and other baselines for cold-start patients across different percentiles, ranked by the number of medical codes, evaluated using PRAUC, F1, Jaccard, and DDI metrics for both MIMIC-III and AKI datasets.}
    
    \label{cold_start}
\end{figure*}

\subsubsection{RQ2: Ablation Studies}
Our proposed approach comprises three main components: Self-Adaptation, Peer-Adaptation, and Uncertainty Filtering. As demonstrated in Tables \ref{tb:ablation_mimic_consistent} and \ref{tb:ablation_aki_consistent}, all of the components complement each other effectively for the drug recommendation task. The No Adaptation baseline only utilizes the patient and visits embeddings with global adaptation during training. The comparison between Self-Adaptation and No Adaptation highlights that adapting the prediction layer using the patient's own visits substantially improves performance. This is particularly important as many patients in the dataset have a limited number of medical codes and visits. Additionally, the comparison between Peer-Adaptation and No Adaptation demonstrates that adapting the model with visits similar to the patient’s last visit enhances the method's performance by directly addressing the issue of visit sparsity. Furthermore, self-adaptation achieves significantly higher performance than peer adaptation, indicating that tailoring the model to the patient’s own visit history provides a more accurate representation of their health profile than relying on data from other patients.  Finally, comparing MetaDrug with MetaDrug w/o UF demonstrates that Uncertainty Filtering supports Self-Adaptation and enhances meta-learning by incorporating only related visits to the patient profile, further boosting overall performance.

\subsubsection{RQ3: Cold-Start Patients}
In this study, we define cold-start patients as those with a limited number of medical codes recorded in their clinical history. Our preliminary analysis revealed that patients with relatively few visits could nonetheless exhibit a high number of medical codes, providing a high precision of the recommendations. In contrast, patients with a limited number of medical codes consistently demonstrated reduced predictive performance, irrespective of the number of visits. These findings motivate our definition, which prioritizes medical code count over visit frequency as the principal criterion for identifying cold-start cases. To apply this definition, we sorted users based on their number of available medical codes for each percentile from 10\% to 50\% to reflect the level of sparsity. As shown in Figure \ref{cold_start}, the proposed approach outperforms all baseline methods across all classification metrics for each percentile. In the 10th percentile, where the number of medical codes is limited, performance is generally lower across methods, and the proposed approach achieves superior results, demonstrating that the two-level adaptation phase effectively addresses the cold-start problem. Another key observation is the difference between MELU and MetaDrug. This comparison highlights that typical item-based adaptation methods used for cold-start users in recommendation systems do not translate well to EHR systems due to structural differences. Furthermore, as the number of medical codes per patient increases, MetaDrug exhibits a greater classification performance advantage over other approaches. This improvement occurs because models have access to more data and potentially more patient visits, allowing for better adaptation. 

\begin{figure}[t]
  \setlength{\abovecaptionskip}{5pt}
  \centering
  \subfloat[MetaDrug]{%
    \includegraphics[width=0.35\textwidth]{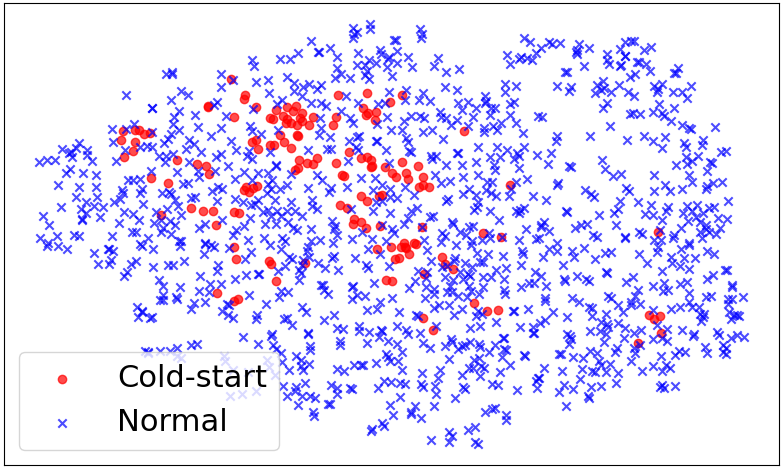}%
    \label{fig:sub1}}
  \hfill
  \subfloat[MetaDrug w/o Adaptations]{%
    \includegraphics[width=0.35\textwidth]{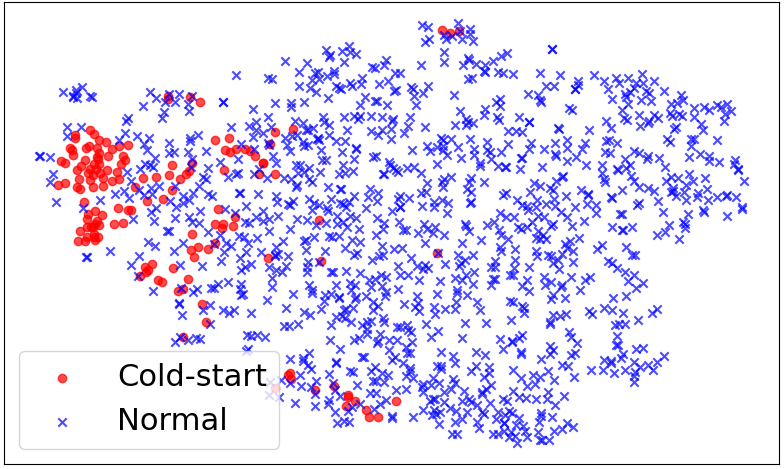}%
    \label{fig:sub2}}
  \caption{t-SNE visualization of the patient embedding $\boldsymbol{E}^p_i$ in the MIMIC-III dataset. Cold-start patients are the lowest 20th percentile ranked by number of medical codes.}
  \label{CLfig}
\end{figure}

Additionally, as shown in Figure \ref{CLfig}, we visualize the t-SNE of the patient embedding dataset for both MetaDrug and MetaDrug without multi-level adaptations. The results indicate that, prior to adaptation, patient embeddings are notably distant from those with a sufficient number of codes. However, after applying adaptation, the method effectively refines the embeddings for cold-start patients, making them more similar to other patients. This demonstrates that the meta-learning approach directly enhances cold-start patient embeddings to be similar to other patients.

\begin{table*}[t]
\centering
\caption{Model selection of various uncertainty quantification methods within the MetaDrug framework on the MIMIC-III dataset, evaluated with 95\% confidence intervals computed across folds.}
\renewcommand{\arraystretch}{1.5}
\setlength{\tabcolsep}{2.5em}
\begin{tabular}{lcccc}
\toprule
\textbf{Models} & \textbf{PRAUC}$\uparrow$ & \textbf{F1}$\uparrow$ & \textbf{Jaccard}$\uparrow$ & \textbf{DDI}$\downarrow$ \\
\midrule
\textbf{Auxiliary Error} & \textbf{0.752} $\pm$ 0.011 & \textbf{0.641} $\pm$ 0.009 & \textbf{0.494} $\pm$ 0.008 & 0.083 $\pm$ 0.004 \\
Deep Ensemble          & 0.746 $\pm$ 0.007 & 0.635 $\pm$ 0.008 & 0.488 $\pm$ 0.005 & 0.082 $\pm$ 0.008 \\
NC Dropout               & 0.748 $\pm$ 0.009 & 0.634 $\pm$ 0.006 & 0.485 $\pm$ 0.007 & \textbf{0.079} $\pm$ 0.008 \\
\bottomrule
\end{tabular}
\label{tb:uq_mimic}
\vspace{-0.3cm}
\end{table*}

\begin{table*}[t]
\centering
\caption{Model selection of various uncertainty quantification methods within the MetaDrug framework on the AKI dataset, evaluated with 95\% confidence intervals computed across folds.}
\renewcommand{\arraystretch}{1.5}
\setlength{\tabcolsep}{2.5em}
\begin{tabular}{lcccc}
\toprule
\textbf{Models} & \textbf{PRAUC}$\uparrow$ & \textbf{F1}$\uparrow$ & \textbf{Jaccard}$\uparrow$ & \textbf{DDI}$\downarrow$ \\
\midrule
\textbf{Auxiliary Error} & \textbf{0.533} $\pm$ 0.005 & \textbf{0.390} $\pm$ 0.004 & \textbf{0.333} $\pm$ 0.003 & \textbf{0.086} $\pm$ 0.005 \\
Deep Ensemble          & 0.529 $\pm$ 0.005 & 0.385 $\pm$ 0.004 & 0.329 $\pm$ 0.004 & 0.091 $\pm$ 0.004 \\
NC Dropout               & 0.531 $\pm$ 0.005 & 0.385 $\pm$ 0.005 & 0.327 $\pm$ 0.004 & 0.088 $\pm$ 0.004 \\
\bottomrule
\end{tabular}
\label{tb:uq_aki}
\vspace{-0.3cm}
\end{table*}

\begin{table*}[t]
\centering
\caption{Hyperparameter selection for different $\beta$ (uncertainty quantification percentiles) on the MIMIC-III dataset, evaluated with 95\% confidence intervals computed across folds.}
\renewcommand{\arraystretch}{1.5}
\setlength{\tabcolsep}{2.5em}
\begin{tabular}{lcccc}
\toprule
\textbf{$\beta$} & \textbf{PRAUC}$\uparrow$ & \textbf{F1}$\uparrow$ & \textbf{Jaccard}$\uparrow$ & \textbf{DDI}$\downarrow$ \\
\midrule
10 & $0.748 \pm 0.008$ & $0.634 \pm 0.007$ & $0.489 \pm 0.006$ & \textbf{0.079} $\pm$ 0.005 \\
\textbf{20} & \textbf{0.752} $\pm$ 0.011 & \textbf{0.641} $\pm$ 0.009 & \textbf{0.494} $\pm$ 0.008 & $0.083 \pm 0.004$ \\
30 & $0.745 \pm 0.010$ & $0.636 \pm 0.007$ & $0.488 \pm 0.005$ & $0.083 \pm 0.006$ \\
\bottomrule
\end{tabular}
\label{tb:beta_mimic}
\vspace{-0.3cm}
\end{table*}

\begin{table*}[t]
\centering
\caption{Hyperparameter selection for different $\beta$ (uncertainty quantification percentiles) on the AKI dataset, evaluated with 95\% confidence intervals computed across folds.}
\renewcommand{\arraystretch}{1.5}
\setlength{\tabcolsep}{2.5em}
\begin{tabular}{lcccc}
\toprule
\textbf{$\beta$} & \textbf{PRAUC}$\uparrow$ & \textbf{F1}$\uparrow$ & \textbf{Jaccard}$\uparrow$ & \textbf{DDI}$\downarrow$ \\
\midrule
10 & $0.529 \pm 0.004$ & $0.383 \pm 0.005$ & $0.328 \pm 0.004$ & $0.080 \pm 0.004$ \\
\textbf{20} & \textbf{0.533} $\pm$ 0.005 & \textbf{0.390} $\pm$ 0.004 & \textbf{0.333} $\pm$ 0.003 & $0.086 \pm 0.005$ \\
30 & $0.526 \pm 0.005$ & $0.385 \pm 0.006$ & $0.329 \pm 0.005$ & \textbf{0.078} $\pm$ 0.004 \\
\bottomrule
\end{tabular}
\label{tb:beta_aki}
\vspace{-0.3cm}
\end{table*}

\subsubsection{RQ4: Model Selection for Uncertainty Quantification}\hspace{5pt}We introduce a model that filters visits based on their predicted uncertainty values. This approach uses black-box post hoc uncertainty quantification methods to estimate the uncertainty associated with each visit for filtering purposes. To this end, we employ three distinct uncertainty quantification models from the prior literature and adapted to the sequential medication recommendation task:
\begin{itemize}[itemsep=0.5ex] 
  \item \textbf{Auxiliary Error Prediction} \cite{thiagarajan2020building} estimates uncertainty by predicting the absolute error between the model's output and the ground truth label. Specifically, the uncertainty value is computed as ${\mathcal{S}}^{(j)}_{i,t_i} = \left| Y^{(j)}_{i,t_i} - \hat{y}^{(j)}_{i,t_i} \right|$. 
  \item \textbf{NC Dropout} \cite{gal2016dropout} uses a Transformer with active dropout (rate 0.5) during inference, generating multiple stochastic forward passes for recommendations. The uncertainty value is computed as the standard deviation of the prediction scores: ${\mathcal{S}}^{(j)}_{i,t_i} = \text{Std}(\hat{y}^{(j)}_{i,t_i,k})$, where $k$ indexes each stochastic forward pass. 
  \item \textbf{Deep Ensemble} \cite{lakshminarayanan2017simple} contains three Transformer models with different random initialization. The uncertainty is quantified as the standard deviation of their predictions: ${\mathcal{S}}^{(j)}_{i,t_i} = \text{Std}(\hat{y}^{m,(j)}_{i,t_i})$, where $m$ denotes the index of the ensemble Transformer model.
\end{itemize}

As shown in Tables \ref{tb:uq_mimic} and \ref{tb:uq_aki}, the proposed approach consistently improves performance across all uncertainty quantification models used in our study. Notably, the Auxiliary Error Prediction model outperforms other methods, indicating that directly learning to predict uncertainty values for filtering yields superior results compared to approaches based on standard deviations of output predictions.

We also conducted an additional ablation study on the parameter $\beta$, percentile for the uncertainty level,  across different percentages. As shown in Tables \ref{tb:beta_mimic} and \ref{tb:beta_aki}, the performance peaks at a $\beta$ value corresponding to 20\%. This suggests that removing too few visits during self-adaptation has a limited impact, while removing too many can hinder the optimization process at this adaptation level.

\section{Related Work}
\textbf{EHR Representation Learning:} In recent years, the emergence of deep learning methods has led to the development of various healthcare-based approaches aimed at capturing the longitudinal attributes of EHR datasets \cite{hadizadeh2024discovering, nayebi2023contrastive}. These methods have been applied in diverse applications, such as medication recommendation \cite{xie2022deep, swinckels2024use} and mortality prediction \cite{choi2020learning}. In the early stages, methods such as Deepr \cite{nguyen2016mathtt}, an end-to-end model that transforms features into sequences of discrete elements and employs a convolutional neural network, Dr. Agent \cite{gao2020dr}, which uses an RNN-based policy gradient for adaptive learning, and RETAIN \cite{retain}, a dual RNN network that incorporates a multilevel attention mechanism for both medical codes and visits, were introduced. While these approaches have enhanced embedding extraction, they do not fully address the challenges of prescription recommendation. In recent years, numerous recommendation systems have been proposed. For instance, GAMENet \cite{gamenet1} leverages a graph-based network to incorporate drug-drug interactions (DDI) as both a loss function and an evaluation metric. Other approaches, such as SafeDrug \cite{safedrug}, use message passing for medication recommendation based on DDI and molecular structures. MICRON \cite{micron} employs a residual neural network to update patient health histories, while MoleRec \cite{molerec} improves drug recommendation safety and performance by considering molecular structures. Additionally, ARCI \cite{ARCI} introduces a multi-level transformer approach to link medications across different visits and intents. Although these methods have shown considerable success in medication recommendation, they remain limited in effectively addressing the cold-start problem in EHR datasets.

\vspace{0.05cm}
\noindent\textbf{Cold-Start Problem in EHR:} The cold-start problem in EHR can be viewed from two perspectives: item cold-start and user cold-start. In the case of the item cold-start problem \cite{song2019medical, shang2019pre}, several approaches have been proposed in recent years to address this issue in EHR data. For example, GRAM \cite{GRAM}, a graph-based attention model, is designed to mitigate data insufficiency while enhancing interpretability. HAP \cite{zhang2020hierarchical} is a medical ontology embedding model that propagates attention across the entire ontology structure, allowing a medical concept to learn its embedding from all concepts in the hierarchy, not just its ancestors. Adore \cite{cheong2023adaptive} introduces an adaptive ontology-integrated representation learning structure that jointly learns category-aware concept embeddings and adjusts ontology relationships via attention to better align with EHR data. Mmore \cite{song2019medical} enables multiple attention-driven representations per medical ontology category to resolve conflicting structure and better align concept embeddings with real-world EHR patterns for improved predictive accuracy. Similarly, KAMPNet \cite{an2023kampnet} is a multi-sourced medical knowledge-augmented medication prediction network that captures diverse relationships between medical codes using a multi-level graph contrastive learning framework. Although these methods have been successful in addressing the item cold-start problem, they do not address the user cold-start problem in healthcare, which remains an unexplored research topic in EHR. In recommendation systems, one common approach to tackling the user cold-start problem is meta-learning \cite{bharadhwaj2019meta}. For example, MELU \cite{MELU} is a meta-learning-based recommender system designed to handle the cold-start problem by rapidly estimating user preferences with only a few consumed items. Similarly, MetaHin \cite{MetaHIN} applies a meta-learning approach to improve cold-start recommendations in heterogeneous information networks (HINs) by integrating meta-learning at the model level and HIN-based semantics at the data level. Additionally, M2EU \cite{M2EU} is a meta-learning method designed to address the cold-start problem by enriching user representations with information from similar users based on attributes and interactions. Although these methods have achieved notable success in medication recommendation, they struggle to address key challenges such as the cold-start problem and the lack of reliable uncertainty quantification for support sets in EHR datasets.
\section{Conclusion}
In this paper, we introduced MetaDrug, a novel multi-level meta-learning framework designed to tackle the patient cold-start problem in medication recommendation systems. Unlike traditional approaches that struggle with patients who have limited historical visits, MetaDrug leverages both self-adaptation and peer-adaptation to enhance predictive accuracy for new patients. Our self-level adaptation enables the model to process multiple medical codes per visit while capturing the sequential dependencies within a patient's history. The peer-level adaptation integrates relevant information from similar visits of other patients. Both adaptation strategies enrich the representation of cold-start patients and mitigate data sparsity issues. Additionally, an uncertainty filtering method is employed during meta-testing to filter out support visits that are not relevant to the patient's profile during the self-adaptation to enhance the generalization of the model during inference. 

\section{AI-Generated Content Acknowledgement}
We utilized generative AI tools (e.g., ChatGPT) exclusively for grammatical and stylistic refinement of the manuscript. These tools were not used for content creation, data analysis, or experimental design.
\bibliographystyle{ieeetr}
\bibliography{references}

\end{document}